\documentclass[conference]{IEEEtran}
\newenvironment{keywords}{
  \noindent \textbf{Keywords:}
  \itshape
}{\par}
\IEEEoverridecommandlockouts

\usepackage{cite}
\usepackage{amsmath,amssymb,amsfonts}
\usepackage{algorithmic}
\usepackage{algorithm}
\usepackage{graphicx}
\usepackage{textcomp}
\usepackage{xcolor}
\usepackage[left=1.62cm,right=1.62cm,top=1.9cm]{geometry}
\def\BibTeX{{\rm B\kern-.05em{\sc i\kern-.025em b}\kern-.08em
    T\kern-.1667em\lower.7ex\hbox{E}\kern-.125emX}}
\begin{document}

\title{Learning Robust Representations for Communications over Noisy Channels\\}

\author{
\IEEEauthorblockN{Sudharsan Senthil\IEEEauthorrefmark{1}, Shubham Paul\IEEEauthorrefmark{1}, Nambi Seshadri\IEEEauthorrefmark{2}\IEEEauthorrefmark{1}, and R. David Koilpillai\IEEEauthorrefmark{1}}
\IEEEauthorblockA{\IEEEauthorrefmark{1}Indian Institute of Technology Madras, Chennai, India \\
Email: \{sudharsansenthil@hotmail.com, paulshubham96@outlook.com, koilpillai@ee.iitm.ac.in\}}
\IEEEauthorblockA{\IEEEauthorrefmark{2}University of California San Diego, USA. Email: naseshadri@ucsd.edu}
}

\maketitle

\begin{abstract}
We explore the use of FCNNs (Fully Connected Neural Networks) for designing end-to-end communication systems without taking any inspiration from existing classical communications models or error control coding. This work relies solely on the tools of information theory and machine learning. We investigate the impact of using various cost functions based on mutual information and pairwise distances between codewords to generate robust representations for transmission under strict power constraints. Additionally, we introduce a novel encoder structure inspired by the Barlow Twins framework. Our results show that iterative training with randomly chosen noise power levels while minimizing block error rate provides the best error performance.
\end{abstract}

\section{Introduction}
\label{sec: Introduction}
This paper considers the design of end-to-end communication systems using deep learning \cite{Kramer_1,Goodfellow_book}. Deep learning has proved to be a powerful framework for analyzing and exploiting complex data patterns and has found wide-ranging applications such as classification, compression, language modelling, language translation, generation of synthetic data, denoising etc.

The design of communication systems for data transmission over noisy channels continues to be an active area of research although many aspects of the problem are well understood. For example, for channels like the band-limited or power-limited additive white Gaussian noise channel, the channel capacity or the Shannon limit which is the maximum data rate (in bits/sec/Hz) for transmission that is achievable while operating arbitrarily close to zero error rate, is well characterized \cite{Gods_paper}. Practical communication systems that perform close to the capacity limit have been designed for various transmission media such as cable, DSL, optical and wireless. These schemes rely upon state-of-the-art error control coding techniques such as Turbo codes\cite{Turbo_code_OG}, LDPC\cite{LDPC_OG}, or Polar codes\cite{Polar_OG} in combination with well-known modulation schemes such as Quadrature amplitude modulation (QAM) or Phase shift keying (PSK). Less well understood in an optimal sense are the impact of non-Gaussian noise, multiplicative fading and the effects of non-linearities in the transmission or receiver chain.

Most deep learning problems are focused on understanding and utilizing the structure of underlying data. In contrast, a deep learning network for reliable communications is required to add structure to the original data which is typically modelled as independent and identically distributed binary random data. This structured  redundancy is necessary to combat impairments in the channel as mentioned previously. When this redundant data is transmitted over a noisy channel, another deep learning network will exploit the redundancy to decode the original data that has been impaired with a low message error rate. Recently there has been significant interest in applying deep learning tools to the design of communication systems \cite{Tim_1, Tim_2, Vishnu_1, Vishnu_2,Pramod_1} as well as using deep learning to build better codes \cite{jiang2019turboautoencoderdeeplearning,Feedback_AE} and use of deep learning to come up with better decoding algorithms \cite{Nachmani_DL_linear} for some of the conventional error correcting codes that don't lend easily to maximum likelihood decoding.

In this work, we focus on the use of autoencoders~\cite{Kramer_1} to create robust latent spaces using only the tools of machine learning and communication and information theoretic bounds without resorting to explicit use of the vast literature on hand-crafted error correcting codes or their decoding methods. A primary reason to undertake this study is that unlike most of the machine learning tasks where the best achievable performance is either not known or good classical approaches are non-existent, here optimal performance is theoretically achievable as well as practical schemes performing close to optimal limits are known and hence serve to benchmark machine learning performance. We study multiple techniques to build deep learning models using the following criteria

\begin{enumerate}
    \item Maximizing the minimum squared Euclidean distance between distinct transmitted codewords.
    \item Minimizing the upper bound on error rate calculated using union bound.
    \item Maximizing the mutual information using the Donsker-Varadhan variational bound.
    \item Use of Parallel encoders to create multiple representations of data and transmit them.
    \item Minimizing KL divergence between  true message and reconstructed message at randomly chosen SNR (Signal-to-Noise Ratio).
\end{enumerate}

\section{End-to-End System Model}
\label{sec: System Model}
In this section, we detail the communication model under study. We also demonstrate how an Autoencoder suits the model. We also discuss the details of the Autoencoder model.

\begin{figure}[htbp]
\centerline{\includegraphics[width=0.9\linewidth]{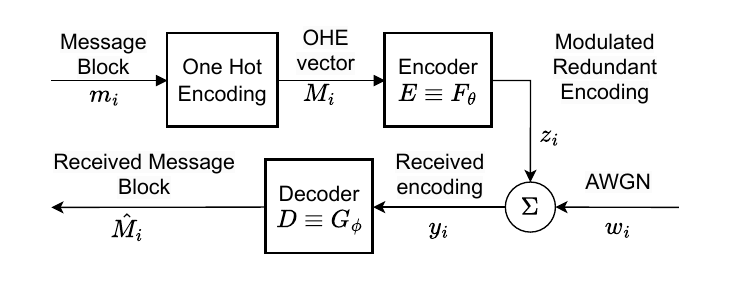}}
\caption{End-to-End System Model}
\label{fig: System model 1}
\end{figure}	

 The diagram of the model is given in Fig~\ref{fig: System model 1}. We intend to transmit a $k$-bit long message $m$ from a transmitter in a noisy channel across $n$ symbols. If $n>k$, we can have some redundancy in transmission and expect it to yield a better BER and BLER performance. Essentially, this mimics the behavior of block codes. We represent a binary message using One Hot Encoding (OHE), so a k-block size binary message will become $2^{k}$-length block vector $M$. The mappings from message $M$ to $n$-symbol codeword are learned by the Fully Connected Neural Network (FCNN)~\eqref{eq: Encoder_eqn}. This can be viewed as a case of coded modulation. An average power constraint is imposed on the encodings produced to mimic the transmit power constraint~\eqref{eq: Power Cons}.

Assume we have a transmitter $Tx$ and a receiver $Rx$. Let $Tx$ transmit the encoder's output $z$. The relation between the input message $M$ and the encoding $z$ is given by \eqref{eq: Encoder_eqn}.
\begin{gather}
E \equiv F_{\theta}: M \rightarrow z 
\label{eq: Encoder_eqn}\\
\mathbb{E}\left[\lVert z^{2} \rVert]\right] = n
\label{eq: Power Cons}
\end{gather}
The encoded vector $z \in \mathbb{R}^n$. Where, $n=k/ r$, is the size of the encoded dimension or the number of symbol transmissions per message block. Throughout this paper, we keep $r=\frac{1}{2}$. The vector $y$ received by $Rx$ is given by \eqref{rx_eqn_1a}.
\begin{gather}
     y = z + w; w \sim \mathcal{N}(0,\sigma^{2}\mathbb{I})
     \label{rx_eqn_1a}\\
      \sigma^{2} = \frac{1}{2rE_b/N_0} \text{, } \label{eq: variance}
\end{gather}
Where $w$ denotes white noise samples, throughout the paper all the experiments are carried out with AWGN channel where the variance is calculated based on~\eqref{eq: variance} from the specified $\frac{E_b}{N_0}$.\\
\begin{gather}
     D \equiv G_{\phi}:  y \rightarrow \hat{M}
     \label{eq: Dec_eq}\\
    P_{e} = \frac{1}{2^k}\sum_{i=1}^{i=2^{k}} P(\hat{M_i}\neq M_i) \label{error_prob}
\end{gather}
The task of the decoder $D$ as given in~\eqref{eq: Dec_eq} is to find the best estimate $\hat{M}$ using the received vector $y$. 

\section{The Autoencoder Model}
\label{The Autoencoder Model}
\begin{figure}[htbp]
\centerline{\includegraphics[width=0.9\linewidth]{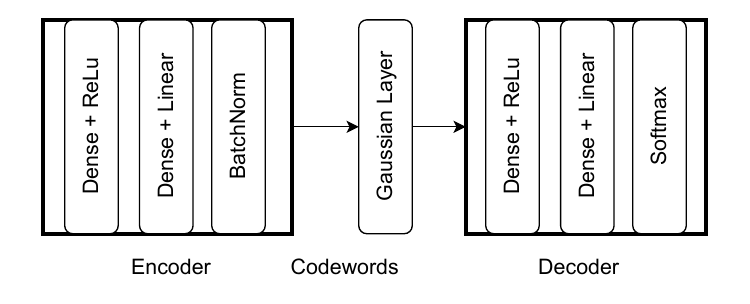}}
\caption{End-to-End Network Architecture}
\label{fig: System model 2}
\end{figure}

The end-to-end network's architecture is given in Fig.~\ref{fig: System model 2}. An optimum decoder minimizes the probability of error in recovering the transmitted messages, given by \eqref{error_prob}. The parameters $\theta$ and $\phi$ associated with the encoder's and decoder's functions are learned during the end-to-end training to optimize cost functions in use. All the models' parameters are learned using Adam optimizer with the default learning rate of 0.001.
As depicted in Fig.~\ref{fig: System model 2}, the encoder has an initial input layer. The input layer accepts one-hot encoded representations of the input message M. This is followed by a hidden Dense layer that is "\textit{ReLu}" activated, which in turn, is followed by a linearly activated layer and a "Batch-Norm" layer. A "Gaussian layer" is inserted to simulate the behaviour of the AWGN channel. The Decoder has one hidden layer which is "\textit{ReLu}" activated followed by a linear layer and an output layer which is "\textit{softmax}" activated. Note that with the softmax layer's output $M^{*}$, $\sum_{j=1}^{j=2^k}M^{*}[j]=1$, where $M^{*}[j]$ denotes the $j^{th}$ element. $\hat{M}$ is the reconstructed message from $M^{*}$ \eqref{eq: Softmax}. Here, $M_j$ refers to the $j^{th}$ message, $j \in [0,2^{k}-1]$. 
\begin{gather}
    \hat{M} = M_{{argmax}_{b} M^{*}[b]}
    \label{eq: Softmax}
\end{gather}
\section{Experiments and Results}
\label{sec: Experiments and Results}
The cost function in \cite{Tim_1} is based on Kullback–Leibler divergence $D_{KL}$ between the true distribution $P_{M}$ and the predicted distribution $Q_{M^{*}}$ for one instance. Note that, $P_{M}(M_b)=0$ $\forall M_b \neq M$ and $Q_{M^{*}}(M_b) = M^{*}[b]$, where $b \in [0,2^{k}-1]$. Eqn~ \eqref{eq:cat_cross_ent} defines the categorical cross-entropy. In \eqref{eq: KL_div} only $H(P_{M},  Q_{M^{*}})$ (9) depends on $\{\phi, \theta\}$ hence \eqref{eq: D_KL_min}. This model results in an end-to-end performance that worsens with the increasing block size, which is undesirable.

\begin{gather}
    D_{KL}(P_{M} \mid\mid Q_{M^{*}}) = \sum_{b=0}^{b=2^{k}-1} P_{M}(M_b) log{(\frac{P_{M}(M_b)}{Q_{M^{*}}(M_b)})}  \label{eq: KL_div}\\ 
   H(P_{M},Q_{M^{*}}) = - \sum_{b=0}^{b=2^{k}-1} P_{M}(M_b) log(Q_{M^{*}}(M_b))  \label{eq:cat_cross_ent}\\  
   \min_{\phi, \theta} D_{KL}(P_{M} \mid\mid Q_{M^{*}}) = \min_{\phi, \theta} H(P_{M},Q_{M^{*}}) 
   \label{eq: D_KL_min} 
\end{gather}
\begin{figure}[htbp]
\centerline{\includegraphics[width=0.8\linewidth]{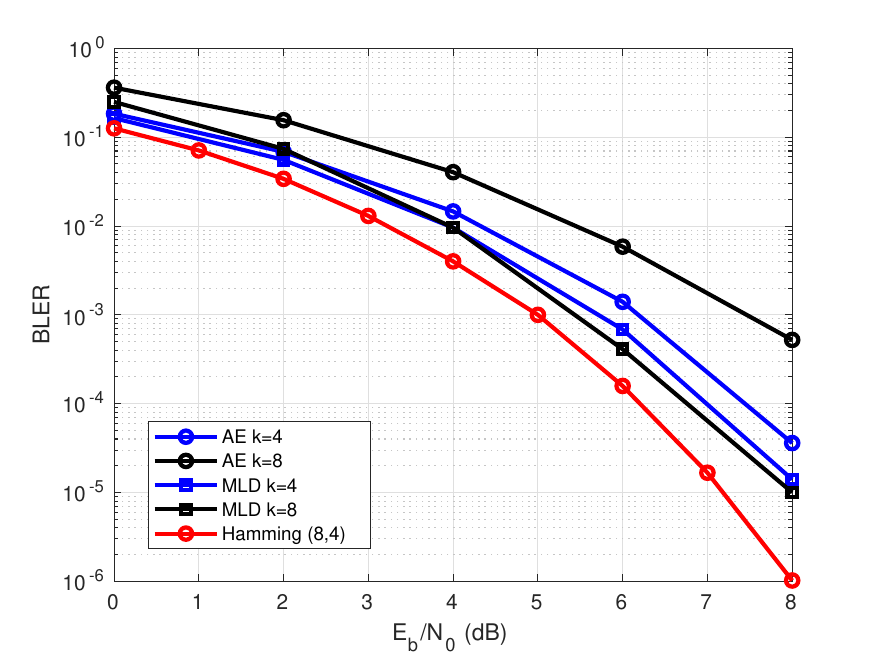}}
\caption{AE models proposed in \cite{Tim_1} vs Maximum-Likelihood Decoders}
\label{fig: AE models proposed by O'Shea vs Maximum-Likelihood Decoders}
\end{figure}
It can be observed from Fig.~\ref{fig: AE models proposed by O'Shea vs Maximum-Likelihood Decoders} that the performance of the model in \cite{Tim_1} falls short of the performance achieved by the neural encoder followed by a  maximum likelihood decoder ~\eqref{eq: Max_Likelihood}. This indicates that the learned neural decoders may not be optimal. Additionally, the performance of the neural encoder with ML-Decoder is not comparable to that of Hamming (8,4), raising questions about the optimality of the encoders. To address this, we suggest novel cost functions for learning the encoder's parameters. Our results demonstrate that models utilizing customized loss functions exhibit significantly enhanced performance. Furthermore, we propose a new encoder structure based on Barlow Twins. Lastly, we introduce a new training strategy that approaches the performance level of Hamming (8,4).
\begin{gather}
    \hat{M_{i}} = {arg max}_{M_{i}} P(y_{i}|M_i)
    \label{eq: Max_Likelihood}\\
    P(y_i|M_i) = P_{y_i|Z=z_i}(y_i|Z=z_i) \sim \mathcal{N}(z_i, \sigma^{2}\mathbb{I})\\
     P_z(z_i) = \frac{1}{M}
    \label{eq: Chn_dist}
\end{gather}

\subsection{Cost Functions for Encoders}
As pointed out in the literature, (\cite{jiang2019turboautoencoderdeeplearning}, \cite{Learn}), straightforward use of FCNNs may not perform better than repetition codes. However, the integration of customized cost functions informed by domain knowledge presents an avenue to guide the model learning process. This approach conduces to a more methodical training methodology, mitigating the risk of optimizers getting ensnared in some local minima. Consequently, the resultant models exhibit enhanced performance when compared to those trained using straightforward methodologies. \\
\textbf{Type I: Maximizing Mutual Information: } The maximization of mutual information between transmitted codewords and received vectors at the receiver (Rx) can reduce the block error rate (BLER) and bit error rate (BER) in end-to-end communication systems, as evidenced in the work by \cite{fritschek2020neuralmutualinformationestimation}. In the study  \cite{fritschek2019deeplearningchannelcoding}, MINE (Mutual Information Neural Estimation) \cite{belghazi2021minemutualinformationneural} was employed to estimate the mutual information between learned codewords and received vectors~\eqref{eq: Information}. However, the applicability of MINE for mutual information estimation diminishes as the block size increases. Additionally, effectively using MINE for mutual information estimation requires extensive hyperparameter tuning and is computationally intensive. Instead, we propose adopting the Weak Law of Large Numbers (WLLN)  \eqref{eq: WLLN} to approximate mutual information, given complete knowledge of the channel's distribution and codewords' distribution \eqref{eq: Chn_dist}. Subsequently, the parameter $\theta$ can be optimized to maximize the $I_{\theta}(Z;Y)$ (I(Z; Y)-estimate).  

\begin{gather}
     I(Z;Y) = \mathbb{E}_{P_{zy}}\left[\log\frac{P_{zy}(z,y)}{P_z(z) P_y(y)}\right]
     \label{eq: Information}\\
    I_{\theta}(Z;Y) = \frac{1}{N} \sum_{i=1}^{i=N} \log\frac{P_{zy}(z_i,y_i)}{P_z(z_i) P_y(y_i)}; I_{\theta}(Z;Y) \approx I(Z;Y)
    \label{eq: WLLN}
\end{gather}
A lower bound on MI can be derived from Donsker-Varadhan's MI variational bound \eqref{eq: DV Lower Bound} by using $T^*$ which is different from the optimal $T$. We substitute the dot-product between codewords and channel output as $T^{*}$ in \eqref{eq: DV Lower Bound}. Similar to using $I_{\theta}(Z;Y)$, $\theta$ can be optimized by maximizing $I^{*}(Z; Y)$ (17-18) which is a lower bound on I(Z; Y).\\
\begin{gather}
    I(Z;Y) = \sup_{T \in \mathcal{F}} \left( \mathbb{E}_{P_{zy}}[T(z,Y)] - \log \mathbb{E}_{P_zP_y}\left[e^{T(z,y)}\right] \right)
    \label{eq: DV Lower Bound}\\
    I^{*}(Z;Y) = \mathbb{E}_{P_{zy}}[T^{*}(z,y)] - \log \mathbb{E}_{P_zP_y}\left[e^{T^{*}(z,y)}\right] 
    \label{eq: DV Lower Bound 2}\\
    I^{*}(Z;Y) < I(Z;Y) \label{eq: DV Lower Bound 3}    
\end{gather}
where $T^{*}(z,y) = z.y$

Steps 5 and 6 in Algorithm~\ref{Training with I(Z; Y)} generate N samples of $z_i$ and $y_i$ to estimate I(Z; Y). In the $2^{nd}$ term in \eqref{eq: DV Lower Bound}, one can mix and match the pairs to have independent samples of $z_i$ and $y_i$. From experiments, we found out that varying a secondary learning factor $F^l$ associated with only MI-maximization in cost functions leads to better performance. We keep  $F^l=100$ and $N=1600$.\\
\textbf{Type II: Maximizing Pairwise Distance: }
In addition to maximizing the MI estimate or MI-lower bound, the encoder's parameters can be optimized by maximizing the pairwise distances between the learned codewords. This can be done by maximizing the sum of squares of pairwise distances \eqref{eq: Pairwise_Dist} and by minimizing the union-bound on error probability \eqref{eq: Union_Bound}. We shall term the former model obtained as \textit{d-Max} model.  More details on the bound \eqref{eq: Union_Bound} can be found in \cite{Gallager_2008}. \\
\begin{gather}
    D_{\theta}(Z) = \sum_{i=1}^{i=2^{k}}\sum_{j=1}^{j=2^{k}}\lVert z_i-z_j \rVert^{2}
    \label{eq: Pairwise_Dist}\\
    U_{\theta}(Z) = \sum_{i=1}^{i=2^{k}}\sum_{j=1}^{j=2^{k}} \exp{\frac{-\lVert z_i-z_j \rVert^{2}}{2\sigma^{2}}}
    \label{eq: Union_Bound}
\end{gather}
\textbf{Type III: Power Constraint: } Along with different combinations of proposed cost functions, a term that enforces unit average power constraint on the codewords is used in place of a batch-normalization layer \eqref{eq: Power_Constraint}. 
\begin{gather}
    P_{\theta}(Z) = \left|\frac{1}{2^k}\sum_{i=1}^{i=2^{k}}\lVert z_{i}^{2} \rVert - n\right|
    \label{eq: Power_Constraint}\\
    L_{\theta,\phi}[P_M,Q_{M^{*}}] =  H(P_{M},Q_{M^{*}})
    \label{eq: L}
\end{gather}

\begin{figure}[htbp]
\centerline{\includegraphics[width=0.8\linewidth]{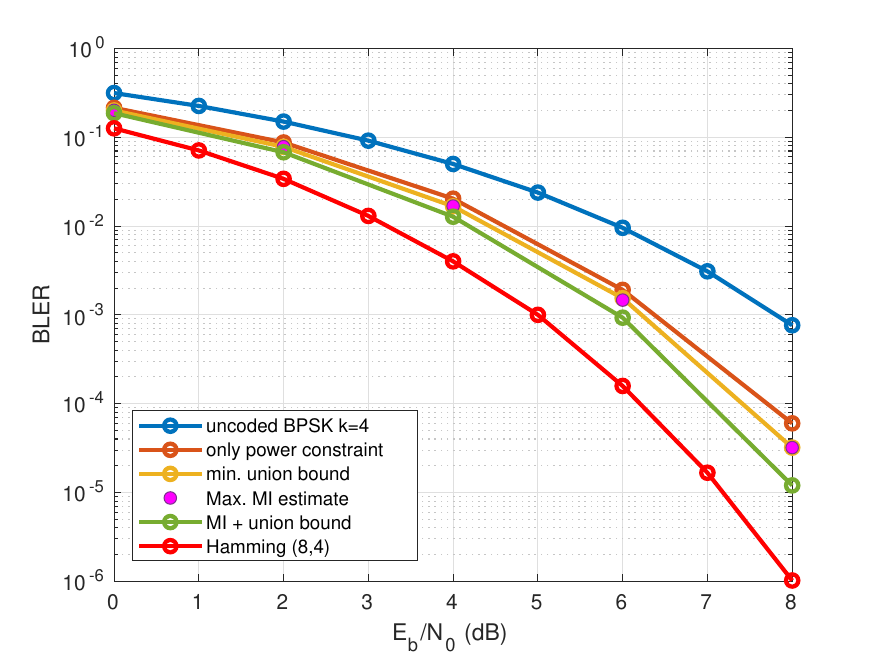}}
\caption{BLER performance with varying Loss Functions and Customised Power Constraints}
\label{fig: Good BLER performance}
\end{figure}

\begin{figure}[htbp]
\centerline{\includegraphics[width=0.8\linewidth]{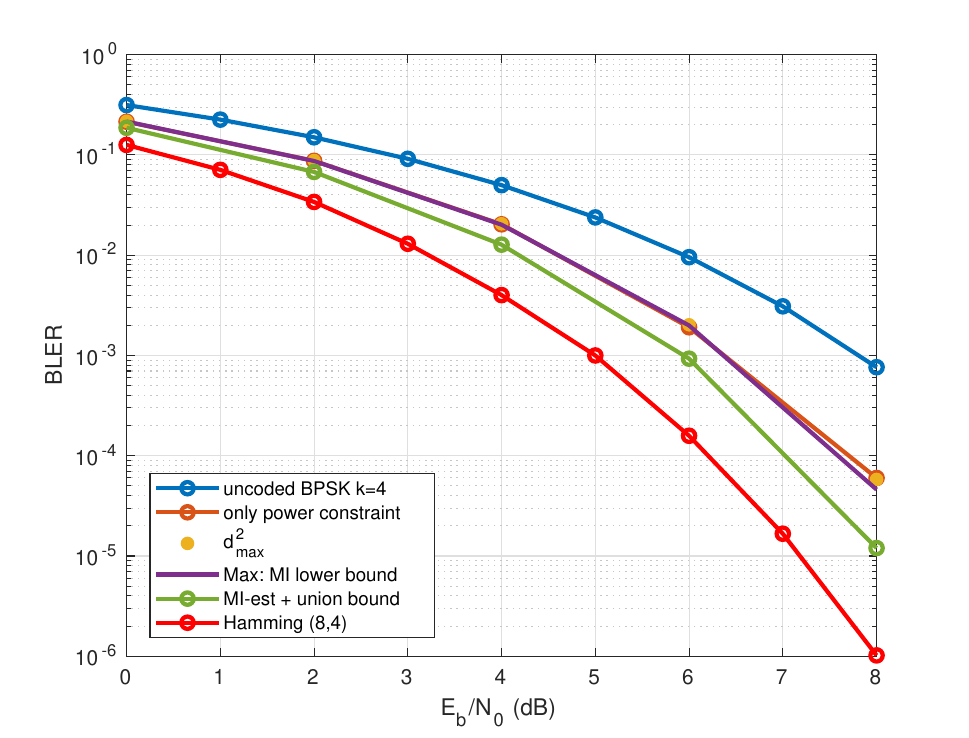}}
\caption{BLER performance with varying Loss Functions which are underperforming in comparison.}
\label{fig: underperforming BLER performance}
\end{figure}

\begin{figure}[htbp]
\centerline{\includegraphics[width=0.8\linewidth]{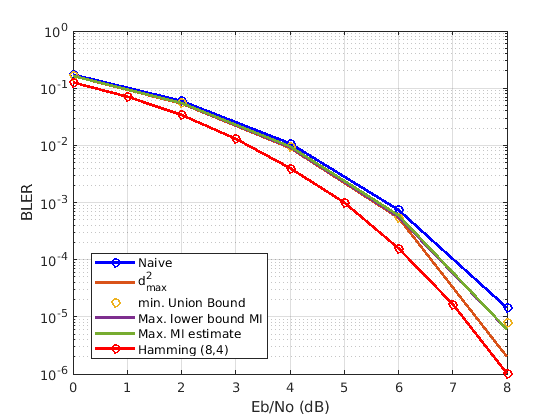}}
\caption{BLER performance for varying cost functions with Batch Norm Power Constraints}
\label{fig: BLER performance for varying cost functions with Batch Norm Power Constraints}
\end{figure}

 In Fig~\ref{fig: Good BLER performance}. we observe the performance obtained upon using union-bound along with an MI estimate this gives the best performance, refer to Algorithm 1. In step 9 $L_{\theta,\phi}[P_M,Q_{M^{*}}]$ comes from ~\eqref{eq: L} . Fig.~\ref{fig: underperforming BLER performance} shows the underperforming cost functions compared to Union bound and MI-estimate. Upon pairing the encoding cost functions with the batch normalization layer, the performance improvement becomes less visible as shown in Fig.~\ref{fig: BLER performance for varying cost functions with Batch Norm Power Constraints}. Algorithm~\ref{Training with I(Z; Y)} contains training with $U(Z)$ and $I_{\theta}(Z; Y)$\ and training with $D(Z)$ and $I^*(Z; Y)$ is very similar and hence not included here. 


\begin{algorithm}[t]
\caption{ Training with I(Z;Y), U(Z) and P(Z)}
    \label{Training with I(Z; Y)}
\begin{algorithmic}[1]
\STATE $\frac{E_b}{N_0} \gets input$ ;
\WHILE{epoch $<$ maxepochs}
    \STATE $z = F_{\theta}(M)$ ; $w \gets AWGN(\sigma)$ ; $y = z + w$
    \FOR{$i<N$}
    \STATE $z_i = F_{\theta}(M_{i\%2^{k}})$ ; $w_i \gets AWGN(\sigma) $
    \STATE $y_i = z_i + w_i$
    \ENDFOR
    \STATE $Y = \{y_1,y_2...y_N\}$ ; $Z = \{z_1,z_2...z_N\}$ ; $M^{*} = D_{\phi}(y)$
    \STATE $C = -L_{\theta,\phi}[M,M^{*}] -U_{\theta} -P_{\theta}(Z) +  F^lI_{\theta}(Z;Y) $ ;
    \STATE $\{\theta,\phi\}_{epoch + 1}  \gets \{\theta,\phi\}_{epoch} + \{\nabla_{\theta}C,\nabla_{\phi}C\}$
\ENDWHILE

\end{algorithmic}
\end{algorithm}

\subsection{Barlow Twin based Encoder Structure}
\begin{figure}[htbp]
\centerline{\includegraphics[width=0.8\linewidth]{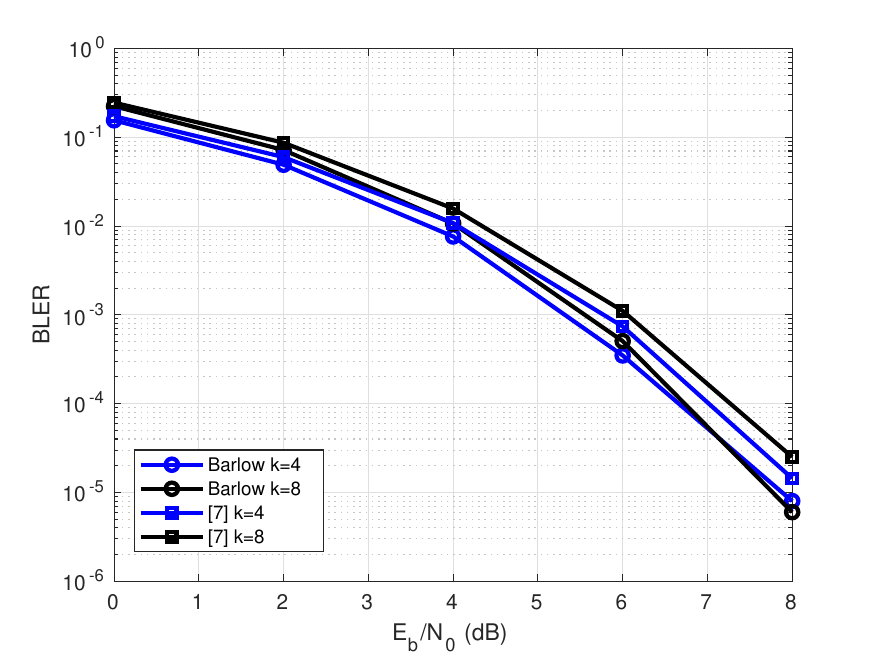}}
\caption{BLER performance of Barlow Twin based structure vs \cite{Tim_1}}
\label{fig: BER performance of Barlow Twins}
\end{figure}

Inspired by \cite{barlow} instead of having one FCNN rate-1/2 encoder, having two parallel rate-1 encoders $F^{1}_{\theta_1}, F^{2}_{\theta_2}$ ~\eqref{eq: Barlow eqn} that are disconnected from each other improves the performance from \cite{Tim_1} Fig.~\ref{fig: BER performance of Barlow Twins}. Note that splitting the encoder decreases the number of learnable parameters. But, naively splitting the encoder any further wouldn't help. In step 1 in Algorithm~\ref{Training Barlow-Twins Network}, z is created by concatenating $z_1$ and $z_2$.

\begin{gather}
    F_{\theta} \equiv F^{1}_{\theta_1}F^{2}_{\theta_2} : M \rightarrow z
    \label{eq: Barlow eqn}
\end{gather}

\begin{algorithm}[t]
    \caption{Training Barlow-Twins Network}
    \label{Training Barlow-Twins Network}
        \begin{algorithmic}[1]
    \STATE $\frac{E_b}{N_0} \gets input$ ;
    \WHILE{epoch $<$ maxepochs}
    \STATE $z_1 = F_{\theta_1}^{1}(M) \text{; } z_2 = F_{\theta_2}^{2}(M)$ ; $z = z_1z_2$
    \STATE  $w \gets AWGN(\sigma)$ ; $y = z + w$
    \STATE  $M^{*} = D_{\phi}(y)\text{; }C = -L_{\theta,\phi}[M,M^{*}]$ ;    
    \STATE  $\{\theta_1,\theta_2,\phi\}  \gets \{\theta_1,\theta_2,\phi\} + \{\nabla_{\theta_1}C,\nabla_{\theta_2}C,\nabla_{\phi}C\}$ 
    \ENDWHILE
\end{algorithmic}
\end{algorithm}

\subsection{Randomized Training SNRs}
\begin{figure}[htbp]
\centerline{\includegraphics[width=0.9\linewidth]{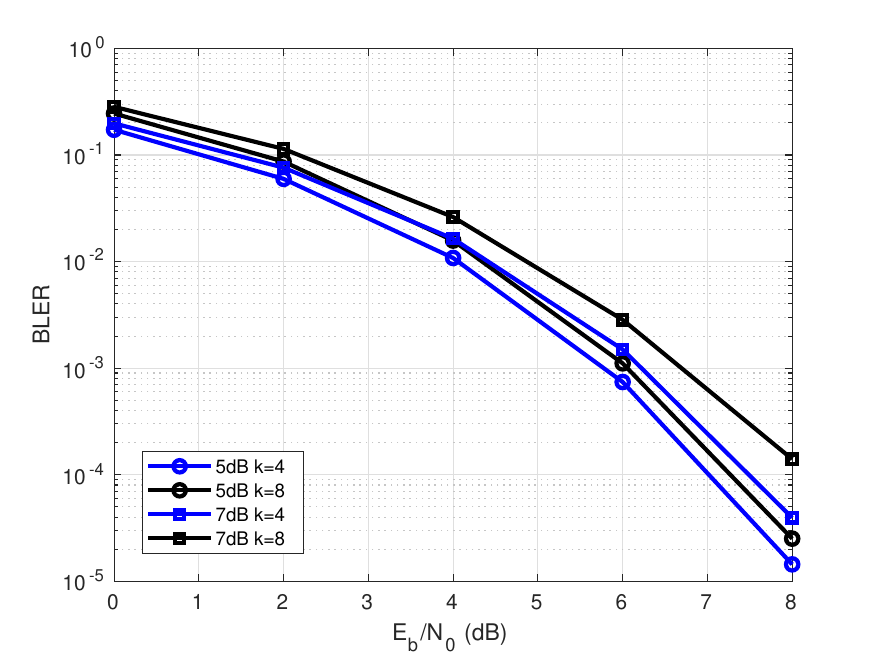}}
\caption{BLER performance of networks trained at 5dB vs 7dB }
\label{fig: BER performance with Training SNRs}
\end{figure}

\begin{figure}[htbp]
\centerline{\includegraphics[width=0.8\linewidth]{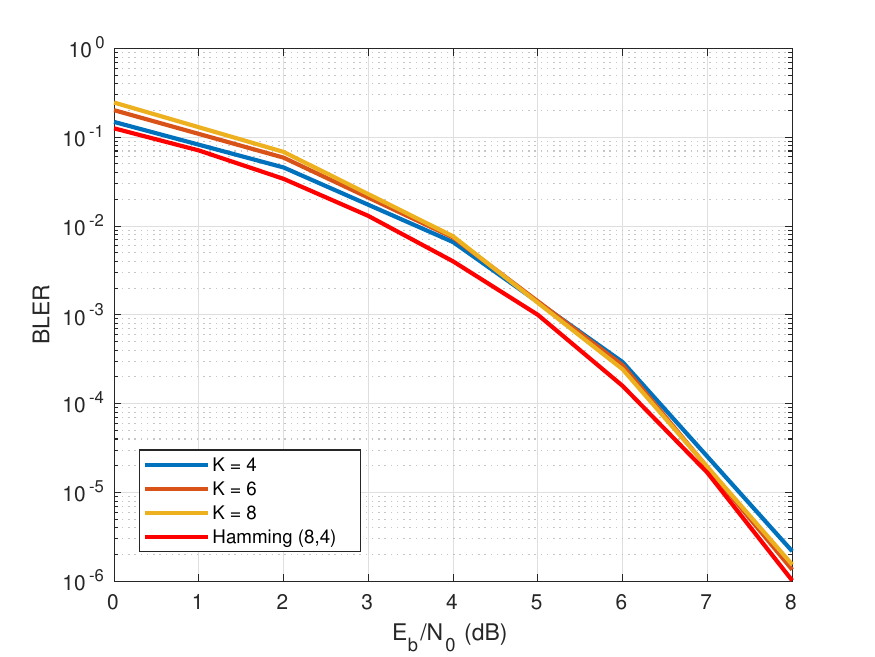}}
\caption{BLER performance of networks trained with random-noise}
\label{fig: BER performance across Block lengths}
\end{figure}

\begin{figure}[htbp]
\centerline{\includegraphics[width=0.8\linewidth]{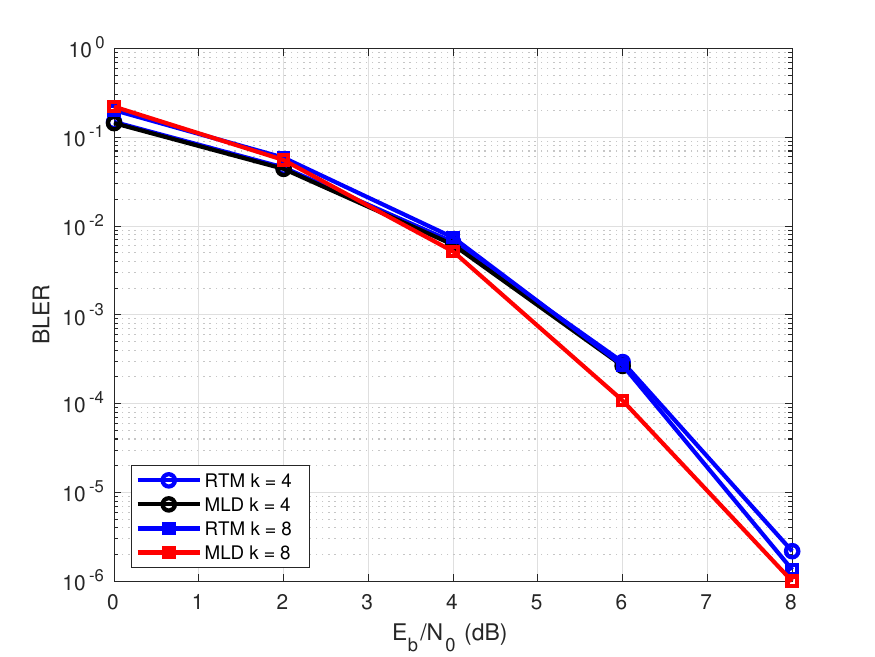}}
\caption{BLER performance of Random-noise trained vs ML-Decoders }
\label{fig: BER performance random vs MLD}
\end{figure}

One can train one end-to-end system for every SNR in a range but this is not feasible, so instead we can train an encoder-decoder with a fixed SNR and evaluate it across a range of SNRs but the performance of these networks is heavily influenced by the training SNR. Fig.~\ref{fig: BER performance with Training SNRs}. This implies that the models fail to generalize well over the "untrained" SNRs. To overcome this problem, we suggest uniformly-randomly sampling from a range of SNRs in our case $\frac {E_{b}}{N_{0}}$ from 0 dB to 12 dB to train the model once and keep doing this iterative until the model converges, refer Algorithm 3. Step 3 in Algorithm 3, $\sigma$ is calculated based on $\frac{E_b}{N_0}$ \eqref{eq: variance}  Fig. \ref{fig: BER performance across Block lengths} presents the best performance so far achieved using only FCNNs to the best of authors' knowledge. We will term this the \textit{RTM}.
Fig.~\ref{fig: BER performance random vs MLD}  shows that randomized training yields improved encoder and decoder combinations. This is evident as the performance is even closer to the ML decoder than the model from \cite{Tim_1}. Note that, Maximum-Likelihood Decoder performance lies exactly on top of Randomized-Noise trained networks for K=4, but it becomes better for K=8. However, there was no visible improvement using randomized noise training with Barlow twins. All the experiments in this subsection are carried out with the same "naive" AE architecture as in Fig.~\ref{fig: System model 2}.\\

\begin{algorithm}
    \caption{Randomized Training}
    \label{alg: Siamese-network}
    \begin{algorithmic}[1]

\WHILE{ epoch $<$ maxepochs}

    \STATE $\frac{E_b}{N_0} \gets uniform[1,12]dB$;  
    \STATE $w \gets AWGN(\sigma(\frac{E_b}{N_0}))$;
    \STATE $z = F_{\theta}(M) ; y = z + w $
    \STATE  $M^{*} = D_{\phi}(y); C = -L_{\theta,\phi}[M,M^{*}]$ ;
    \STATE  $\{\theta,\phi\}_{epoch+1}  \gets \{\theta,\phi\}_{epoch} + \{\nabla_{\theta}C,\nabla_{\phi}C\}$

\ENDWHILE
\end{algorithmic}
\end{algorithm}

\section{Latent Space or Codeword Analysis}
\begin{figure}[htbp]
\centerline{\includegraphics[width=0.8\linewidth]{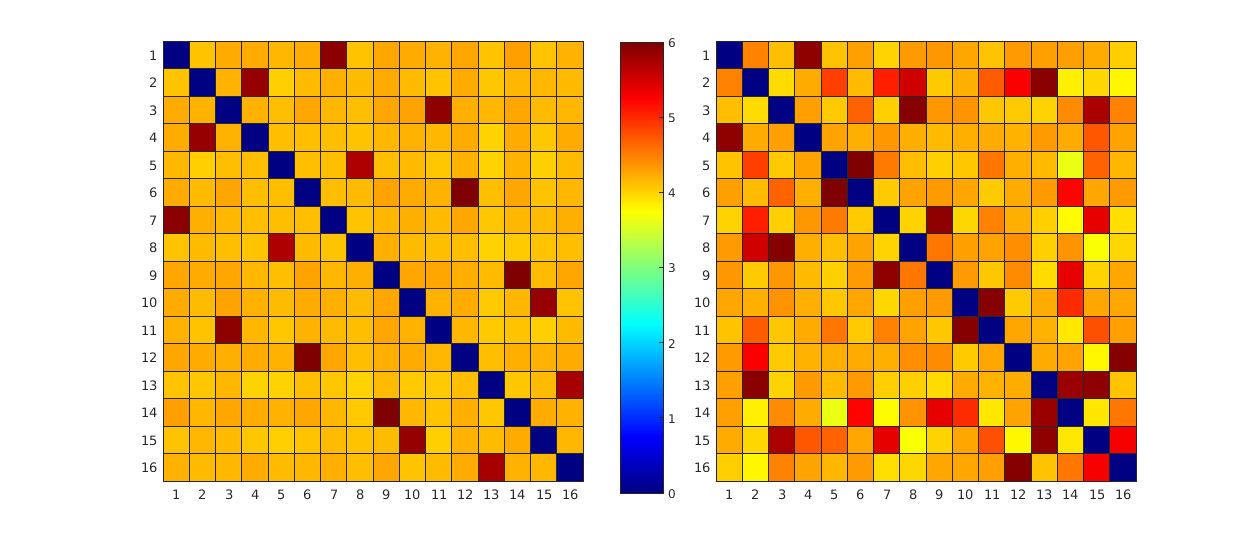}}
\caption{Pairwise mutual distance between codewords produced by the AE-based models. (a): \textit{RTM} (b): \textit{d-Max} model}
\label{fig: Pairwise Mutual Distance}
\end{figure}

Autoencoders generate transmitted codewords in their latent space, as illustrated in Fig.~\ref{fig: System model 2}. In this section, we analyze the properties of the latent space. As discussed in the preceding subsections, the decoder's performance is significantly influenced by the obtained codewords. The n-dimensional codewords reside on the surface of an n-dimensional hyper-surface. Effective decoder performance is achieved when the distance between codewords is maximized, as smaller distances increase the susceptibility to errors caused by noise. 
Practical codewords for a power-limited transmitter are obtained by imposing power constraints, as detailed in the preceding section. We analyze and compare codewords from the two models with the best BLER performance. Fig.~\ref{fig: Pairwise Mutual Distance}-a and b illustrate the pairwise mutual distance of codewords for the \textit{RTM} and $d_{max}^2$ models respectively. As noted in the previous section, the randomized network exhibited the best BLER performance.

\begin{table}
    \centering
    \begin{tabular}{|c|c|c|c|c|c|l|} \hline 
         &  $d_{min}$&  $N_{d_{min}}$&  $d_{avg}$&  $N_{d_{avg}}$& $d_{max}$&$N_{d_{avg}}$\\ \hline 
         RTM&  3.97&  4&  3.99&  2&  5.98&2\\ \hline 
         d-Max model&  3.46&  2&  3.97&  2&  5.75&2\\ \hline
    \end{tabular}
    \caption{Pairwise minimum distance}
    \label{tab: min-dist}
\end{table}

Analyzing Fig.~\ref{fig: Pairwise Mutual Distance}, we observe that randomized training leads to nearly equidistant codewords on the hypersphere. In contrast, the $d_{max}^2$ model achieves varying levels of separation between codewords, resulting in inconsistent performance. Table~\ref{tab: min-dist} details the distance properties. $D_{min}$ is the minimum, $D_{avg}$ is the average and $D_{max}$ is the maximum pairwise distance between codewords. And, $N_{D_{min}}$ gives the number of codewords at $D_{min}$ and so on. From Table~\ref{tab: min-dist} we observe that the RTM model has a larger minimum distance than the $d_{max}^2$ model. And, the $d_{max}^2$ model has a lesser number of codewords that lie on the said minimum distance than the RTM model. Further study of this is deferred to future work.

\section{Conclusions and Future Work}
\label{sec: Conclusion}
In this work, we have demonstrated that FCNNs can be effectively utilized to learn an end-to-end redundant communication system without any inspiration from existing classical Encoder-Decoder algorithms. We have studied the impact of varying cost functions on improving codewords performance with hard power constraints. We have also demonstrated a new encoder structure based on the Barlow-twins philosophy to build a less complex and better-performing encoder. In the end, we show the performance achieved with randomized training is the best so far and is very close to the ML-Decoder's performance on the learned codewords.

Future works in this domain will study the larger block lengths. We would also like to leverage this work to move to cases that study MIMO, and fading channels and further extend to interference-limited channels.
\bibliographystyle{IEEEtran}
\bibliography{main}

\end{document}